\pdfoutput=1
\documentclass{article} 
\usepackage{nips12submit_e,times}

\usepackage{array}
\usepackage{relsize}
\usepackage{multirow}
\usepackage{amsmath}
\usepackage{amsfonts, bm}
\usepackage{amsthm}
\usepackage{graphicx}
\usepackage{wrapfig}
\usepackage{enumerate}
\usepackage{color}
\usepackage[numbers]{natbib}
\usepackage[font=small,format=plain,labelfont={bf,up}]{caption,subfig}
\usepackage{xspace}
\usepackage[normalem]{ulem}

\newcommand{\bpi}{\boldsymbol\pi}
\newcommand{\bgamma}{\boldsymbol\gamma}
\newcommand{\mmm}{$\text{M}^3$\xspace}

 \belowdisplayskip \abovedisplayskip

\newenvironment{packed_enum}{
\begin{enumerate}
  \setlength{\itemsep}{0pt}
  \setlength{\parskip}{0pt}
  \setlength{\parsep}{0pt}
}
{\end{enumerate}}

\newlength{\sectionReduceTop}
\newlength{\sectionReduceBot}
\newlength{\subsectionReduceTop}
\newlength{\subsectionReduceBot}
\newlength{\abstractReduceTop}
\newlength{\abstractReduceBot}
\newlength{\captionReduceTop}
\newlength{\captionReduceBot}
\newlength{\subsubsectionReduceTop}
\newlength{\subsubsectionReduceBot}

\newlength{\horSkip}
\newlength{\verSkip}

\newlength{\figureHeight}
\title{Multidimensional Membership Mixture Models}

\author{
Yun Jiang, 
Marcus Lim
and Ashutosh Saxena\\
Department of Computer Science\\
Cornell University\\
Ithaca, NY 14850 \\
\texttt{\{yunjiang, mkl65, asaxena\}@cs.cornell.edu} \\
}

\nipsfinalcopy 

\begin{document}

\maketitle


\begin{abstract}
\vspace*{\abstractReduceBot}

We present the multidimensional membership mixture (\mmm) models where every dimension of the
membership represents an independent mixture model and each data point is 
generated from
the selected mixture components jointly. This is helpful when the data has a certain 
shared structure. For example, three unique means and three unique variances
can effectively form a Gaussian mixture model with nine components, while
requiring only six parameters to fully describe it.  
In this paper, we present three instantiations of \mmm models (together with
the learning and inference algorithms): 
infinite, finite, and hybrid, depending on whether the number of mixtures is fixed or not.
They are built upon Dirichlet process mixture models, latent Dirichlet allocation,
and a combination respectively.
We then consider two applications: topic modeling and learning 3D 
object arrangements. Our 
experiments show that our \mmm models achieve better performance 
using fewer topics
than many classic topic models. We also observe that topics from the 
different dimensions of \mmm models are meaningful and orthogonal to each other. 

\end{abstract}


\section{Introduction}
\vspace*{\sectionReduceBot}


Inherited from the mixture models' ability to approach complicated distribution using a small set of
simpler distributions, topic models are used to capture `topics'---distributions over the
vocabulary---shared across different documents~\cite{deerwester1990indexing,LDA,HDP}. This concept
has also been successfully applied to 
building image hierarchy~\cite{cao2007spatially,li2010building,sudderth2006describing,
LiSocherFeiFei2009}, where feature-object-scene relationships
follow the word-topic-document analog.

\begin{wrapfigure}[14]{r}{0.25\linewidth}
\vskip -.2in
\centering
\includegraphics[width=1.0\linewidth,clip=true,trim=41 42 0 0]{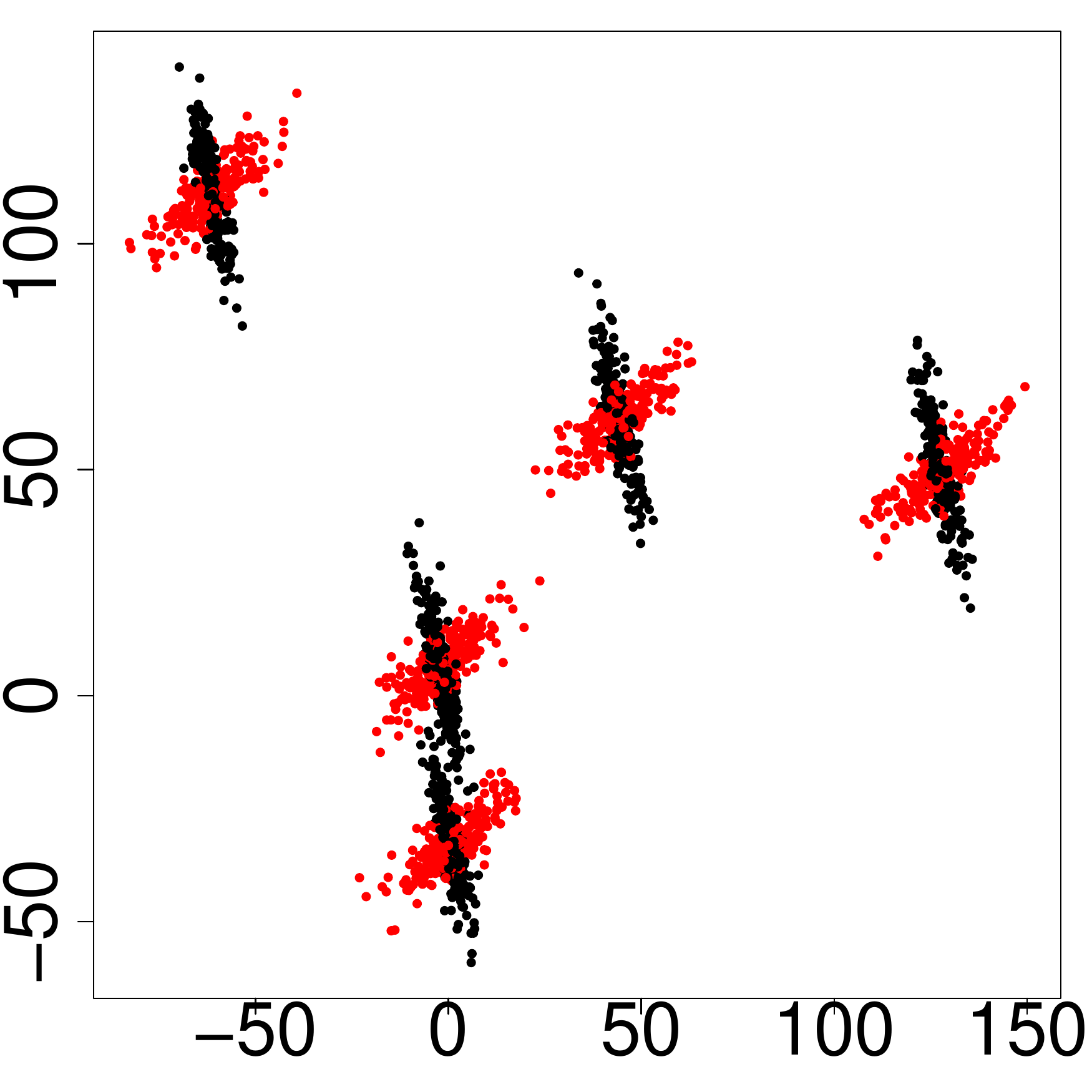}
\vskip -.1in
\caption{A mixture of ten Gaussians, with five unique means and two unique
covariance matrices.}\label{fig:gaussian_intro}
\end{wrapfigure}

Most previous models consider that a word is generated from one type of topic (which we call 
a single-dimensional membership). However, in some cases, an observation may be generated from two or
multiple \emph{types} of topics. 
As a pedagogical example, let us consider a case of Gaussian mixture modeling. 
Fig.~\ref{fig:gaussian_intro} shows a dataset generated from 10 different Gaussian distributions, 
while we only need five \emph{unique} means and two \emph{unique} covariances.
Unless we capture this effect---that the data is generated by two components from different
parameter spaces---our model would not be parsimonious and thus susceptible to over-fitting. 
Such a scenario also happens in document modeling.

In this work, we present multidimensional membership mixture (\mmm) models 
 in which every data point has a multidimensional membership. Each dimension is a 
draw from an independent mixture model. As for the example above, we can use one mixture model for
the means and another for the covariances. As a result, each point has a 2D membership.  
This  leads to parsimonious representations when the data has a certain shared structure. 

Now let us take topic modeling for the NIPS corpus$^{\ref{nips-corpus}}$ as an example. 
Topics that the words are drawn from are the result of combining common topics 
(e.g., `algorithm' and `result') and session-specific topics (e.g., `neural' and `control').\footnote{For simplicity, we use one keyword to represent a topic here.}
Those topics are orthogonal in the sense of that papers of either neuroscience or control theory 
would have content about methods and results. Using \mmm models 
can not only obviate the need of unnecessary topics, but also 
identify such orthogonal structures in the data.

We can also view this multidimensional membership as a way of sharing parameters---observations
assigned to different mixture components in one dimension may be assigned to the same component in
another dimension and thus effectively share the parameters.
This is different from hierarchical topic models,
such as hierarchical DP~\cite{griffiths2004hierarchical,HDP,focussedTopicModelsNIPSws}. Although they allow sharing among components branching from the same
ancestors, the total number of components needed to model the data is not reduced accordingly.

Similar to other topic/mixture models, \mmm models need to infer latent membership for each data
point. But the coupling of multiple mixture models through observed data makes inference more
challenging. In this paper, we formulate and derive three instantiations of \mmm models:
infinite \mmm, finite \mmm and hybrid \mmm, according to whether the number of mixture 
components is fixed or not. They are built upon multiple Dirichlet process mixture models (DPMMs), 
multiple latent Dirichlet allocation (LDA) models, and a combined DPMM and finite mixture model.

In the experiments, we first present a proof-of-concept example of applying the infinite \mmm model on the task
of Gaussian mixture modeling.
We then consider the task of document modeling. 
We compare our finite \mmm model with different topic models (including LDA, hierarchical DP, etc.) on
several corpora. Extensive experiments show that our model achieves lower perplexity on the hold-out
documents while having 
fewer topics than the baselines. The extracted topics from different the dimensions
also reflect certain orthogonality. 

Finally, we also consider the problem of learning 3D object arrangements (which is
useful in scene understanding, object recognition and robotic applications).
An object being at a particular place is governed by two orthogonal factors---its affordances (i.e., how it is used by humans such as
drinking or touching)~\cite{montesano2008learning} and potential human poses (e.g., sitting in a chair or browsing a
book shelf)~\cite{Gupta2011,Grabner2011}. Therefore we use 
two independent mixture models and enable objects of different usages be associated with one
human pose and vice versa. Results show that our hybrid \mmm model outperforms both finite and
infinite mixture models whose mixture components are defined on the joint space of human poses and
object affordances.

\vspace*{\sectionReduceTop}
\section{Related Work}
\vspace*{\sectionReduceBot}

There is a huge body of work employing mixture models. Here we 
only name a few in the area of probabilistic topic models. 
More recent developments in topic modeling can be found in~\cite{blei2011introduction,steyvers2007probabilistic}.

Most methods used to extract topics from a document corpus are grounded in latent variable models
and statistical decomposition techniques,
such as mixture of unigrams model~\cite{nigam1999using}
and probabilistic latent semantic indexing
model (pLSI)~\cite{hofmann1999probabilistic}. Later, Blei et al.~\cite{LDA} proposed LDA to model the
hierarchy of a corpus to
allow different documents to share similar topic proportions and words from one document are
sampled from the same topic distribution. It is later extended to nonparametric hierarchical
models~\cite{griffiths2004hierarchical,HDP,focussedTopicModelsNIPSws}, so that the hierarchy and the number of topics are
learned together. 
The inferential difficulty in those topic models can be alleviated by 
using variational inference~\cite{minka2002expectation,LDA,teh2007collapsed} or MCMC
sampling~\cite{neal2000markov},
including collapsed Gibbs sampling~\cite{griffiths2004finding}. 
These models consider single-dimensional topics, and therefore are complementary our method.

Many models relax the assumptions in LDA by
 modeling word non-exchangeability~\cite{wallach2006topic,griffiths2005integrating}, or by modeling the
correlations among topics~\cite{blei2007correlated, NIPS2011_1117, Putthividhya:2009:IFT:1553374.1553481}. These ideas are complementary to ours, and similar techniques may  
be applied to \mmm models.  Further, there has also been work on incorporating other meta-data such as
authors~\cite{author-topic,dai2011grouped}, citations~\cite{Nallapati:2008:JLT:1401890.1401957}, and
tags~\cite{Das:2011:SJC:2063576.2063773}. Our \mmm model does not require such meta-data.
More importantly, none of these extensions consider the factorization of topics into multiple mixture models.


Topic model have been widely applied to computer vision applications such as
building image hierarchy~\cite{li2010building}, object detection~\cite{sudderth2006describing}, activity
recognition~\cite{wang2009unsupervised}, classification, annotation and segmentation~\cite{LiSocherFeiFei2009}. In some applications, the model is augmented with spatial information to 
yield spatially coherent topics~\cite{wangbleifeifei08}. However, none of the models
presented in these works consider generation of data-points as a multi-dimensional mixture
of topics.

There are previous works in matrix factorization \cite{ding2008equivalence}, factored models
\cite{RAN10}
and parameter sharing \cite{kim2010tree,jalalidirty,li2011_thetamrf,Mei:2008:TMN:1367497.1367512,nicta_4807},
where a lower dimensional representation of the parameters is used. Even though 
these approaches are quite different from our \mmm models, they are relevant to our work since \mmm
model also uses a compact ``factored'' representation for the parameters.


Our model does not diverse far from multidimensional
clustering~\cite{multidimensional-clustering}, two-way
groupings~\cite{twowaygroupings2,twosidedclustering} and some biclustering
models~\cite{madeira2004biclustering}. However in this work, we are  interested in 
modeling the posterior density and topics instead of clustering. 

Many collaborative filtering methods are also built upon mixture models~\cite{jin2006study,marlin2004multiple}, where user
preferences are often modeled by different mixture components. This is similar to classic topic
models whose membership is single-dimensional. 
The flexible mixture model~\cite{fmm} however considers 2D
membership (user and object group) for an observed rating score. It is close to our work but we
consider (Dirichlet) priors and infinite number of groups.

\vspace*{\sectionReduceTop}
\section{Multidimensional Membership Mixture (\mmm) Models\label{sec:cmm}}
\vspace*{\sectionReduceBot}

We present the general idea of our approach in this section, and then describe three specific
instantiations in the next section.

A mixture model typically consists of $K$ mixture components, 
each of which is a distribution parameterized by $\theta_k$, denoted as $F(\theta_k)$.  
Drawing a data point
$x$ involves choosing a mixture component $z \in \{1,\ldots,K\}$ 
according to the mixing proportions $\bpi=(\pi_1,\ldots,\pi_K)$ (subject to $\sum_{k=1}^K
\pi_k = 1$), and then drawing from $F(\theta_z)$, i.e.,
\begin{equation}
z |\boldsymbol \pi \sim \boldsymbol \pi;
\quad\quad\quad\quad\quad\quad
x| z, \theta \sim F(\theta_{z}).
\end{equation}
Depending on whether $K$ is fixed or not, mixture models can be categorized as finite
and infinite (or nonparametric) mixture models. 

Our \mmm models assume data is generated \emph{jointly} by several 
independent mixture models. Particularly, an L-dimensional \mmm model of $L$ different mixture models, each
having $K^\ell$ components parameterized by $\theta^{\ell}_k$. Now, generating a data point $x$ involves  
choosing a mixture component $z^{\ell} \in \{1,\ldots,K^{\ell}\}$ for \emph{each} of the $L$ dimensions.
Given $\Theta = (\theta^\ell_k)^{\ell=1\ldots L}_{k=1\ldots K^\ell}$ and $\boldsymbol
z=(z^1,\ldots,z^L)$,
we then draw $x$ from the distribution parameterized by the selected $L$ mixture components together:
\begin{equation}
z^{\ell} |\boldsymbol \pi^{\ell} \sim \boldsymbol \pi^{\ell},\quad \forall \ell=1,\ldots,L;
\quad\quad\quad\quad\quad
x|\boldsymbol z, \Theta \sim F(\theta^1_{z^1},\ldots,\theta^L_{z^L}).
\end{equation}
Note that the domain of the density function $F$ is now a Cartesian product of the domains of $L$
mixture models, which may or may not be the same. 

\mmm models are related to the standard mixture models in the following ways. When $L=1$, it degenerates 
to the standard mixture model. When $L>1$, we can also cast it into 
an equivalent (single-dimensional) mixture model by defining a new mixture
component for any combination of $L$ components as $\theta'_k=(\theta^1_{j_1},\ldots,\theta^L_{j_L})$
where $j_{\ell}\in \{1,\ldots,K^{\ell}\}$. This leads to a total of $\prod_{\ell=1}^L K^{\ell}$ mixture
components. When $L$ or $K^l$ is large, the corresponding mixture model would be prohibitive to
compute and may tend to over-fit the data. 
On the other hand, \mmm models only construct $\sum_{l=1}^L K^\ell$ mixture components. While this is much more
parsimonious, our method relies on the assumption that the data 
is drawn from shared mixture components whose parameters are generated from independent processes.
Our model would also be able to obtain better estimates of the parameters because 
now more observations would effectively be used for computation.




\smallskip
Note that an $L$-dimensional \mmm model is not the same as $L$ independent mixture models, 
as they are \emph{linked through the observations}. This coupling would result in challenges in 
the inference, such as when optimizing parameters of the $L$ mixture models jointly or sampling from their 
joint posterior distribution.  In the following sections, we will show how to derive and use some
specific \mmm models. 

\begin{figure}[!t]
\vskip -.2in
\centering
\subfloat[DPMM (same as 1D Infinite \mmm)]{
\includegraphics[height=.22\linewidth]{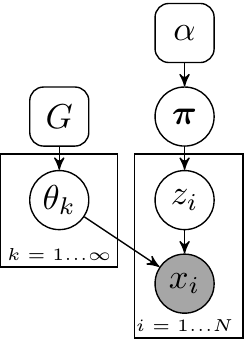}
\quad
}
\subfloat[2D Infinite \mmm model]{
\includegraphics[height=.22\linewidth]{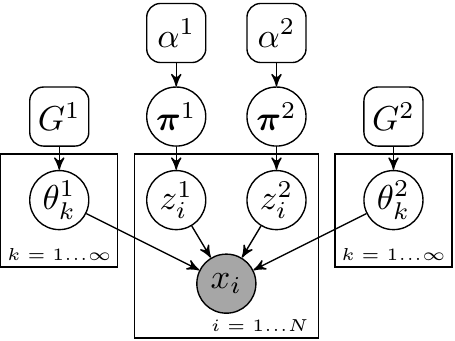}}
\quad\quad
\subfloat[LDA  (same as 1D Finite \mmm)]{
\includegraphics[height=.22\linewidth]{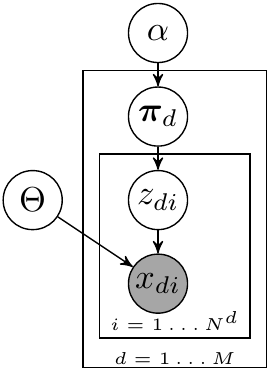}}
\quad
\subfloat[2D Finite \mmm model]{
\includegraphics[height=.22\linewidth]{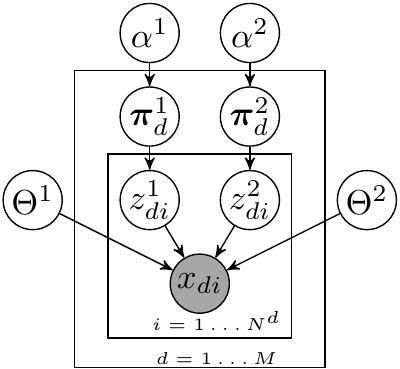}}
\vspace*{\captionReduceTop}
\caption{Two instantiations of our \mmm models: infinite and finite 2D \mmm models and their
corresponding models in 1D.} 
\vspace*{\captionReduceBot}
\label{fig:models}
\end{figure}

\vspace*{\sectionReduceTop}
\section{Formulation and Inference for Three Instantiations of \mmm models}
\vspace*{\sectionReduceBot}
In this section, we describe our three specific instantiations of the
\mmm model. Each is a combination of $L=2$ mixture models, and we informally
call them 2-D \mmm.\footnote{Although we only present 2-D cases, generalization to $L>2$ is straightforward.}
Experiments on each of these models will be presented in Section~\ref{sec:experiments}.

\vspace*{\subsectionReduceTop}
\subsection{Infinite $M^3$ Models using Dirichlet Processes}\label{sec:infinite}
\vspace*{\subsectionReduceBot}
When the number of components, $K$, is unknown, nonparametric
Bayesian methods are often used. For example, Dirichlet process mixture model (DPMM), which is also referred as infinite
mixture model, can adapt $K$ to the data
automatically (overview of DP can be found 
in~\cite{Teh2010}). 
DPMM can be constructed
using a stick-breaking process: 
\begin{equation}\label{eq:dp_pi}
\bpi \sim \text{GEM}(1,\alpha) \quad\quad\quad
\theta_k \sim G_0 
\quad\quad\quad 
z_i |\boldsymbol\pi \sim \boldsymbol\pi\quad\quad\quad
x_i | z_i,\theta_{1:\infty} \sim F(\theta_{z_i})
\end{equation}
where $G_0$ is the base distribution of $\theta$ and $\alpha$ is the concentration parameter. 
Chinese restaurant process provides another perspective to understand how $z_i$ is selected:
\begin{equation}
z_i = z | \boldsymbol z^{-i} = 
\begin{cases}
\frac{n_z^{-i}}{N-1+\alpha} & \text{if $z$ is previously used}\\
\frac{\alpha}{N-1+\alpha} & \text{otherwise}
\end{cases}\label{eq:dp_CRP}
\end{equation}
where superscript $-i$ denotes everything except the $i^{th}$ instance and $n_z^{-i}$ equals the number
of data points assigned to the component $z$ excluding $x_i$.

We formulate the 2-D infinite \mmm model as a combination of two DPMMs, as shown in Fig.~\ref{fig:models}b. Each DP mixture model
follows the same stick-breaking process as in Eq.~\eqref{eq:dp_pi}, except that $x_i$ is now sampled as
\begin{equation}
x_i |z_i^1, z_i^2, \theta^1_{1:\infty}, \theta^2_{1:\infty} \sim F(\theta^1_{z_i^1}, \theta^2_{z_i^2}).
\end{equation}  
We now define the conditional distribution for $z_i^1, z_i^2$, the counterpart of
Eq.~\eqref{eq:dp_CRP} in our \mmm model. 
Let $n^{-i}_{cd}$ equals the number of observations (excluding $x_i$) with $z^1_j = c$ and $z^2_j =
d$. And let $n^{-i}_{c\cdot}=\sum_d n^{-i}_{cd}$ and $n^{-i}_{\cdot d}=\sum_c n^{-i}_{cd}$.
If we assume $z^1_i$ and $z^2_j$ are independent, the joint conditional is decomposed
into their own conditional same as Eq.~\eqref{eq:dp_CRP} by replacing $n_z^{-i}$ with
$n_{z\cdot}^{-i}$ and $n_{\cdot z}^{-i}$.
However, this is such a strong assumption that does not hold in general. 
Therefore, we introduce a \textit{sharing parameter} $\omega \in [0,1]$ to 
control the correlation between the two. 
We define the joint conditional as follows,\footnote{Note that the superscript in the symbols $z$, $\omega$, etc.~denotes the dimension ($\ell=1,\ldots,L$), not the exponent.}
\begin{equation}\label{eq:dp2_CRP}
z^1_i = c, z^2_i = d | \boldsymbol z^{1,-i}, \boldsymbol z^{2,-i} =  
\begin{cases}
\frac{(1-\omega) n^{-i}_{c\cdot} n^{-i}_{\cdot d} + \omega n^{-i}_{cd} (N-1)}{(N-1)(N-1+\alpha)} & \text{$n^{-i}_{c\cdot}>0$ and $n^{-i}_{\cdot d}>0$}\\
\frac{\omega^1\;\alpha\; n^{-i}_{\cdot d}}{(N-1)(N-1+\alpha)} & n^{-i}_{\cdot d}>0\\
\frac{\omega^2\;\alpha\; n^{-i}_{c \cdot}}{(N-1)(N-1+\alpha)} & n^{-i}_{c\cdot}>0\\
\frac{\omega \alpha}{N-1+\alpha} & \text{otherwise}
\end{cases}
\end{equation}
with 
$\omega^1$ and  $\omega^2$ (subject to $\omega+\omega^1+\omega^2 = 1$)  used to tune the relative concentration parameter between the two
dimensions. Under this definition,  $\omega$ constructs a smooth continuum between 
2-D \mmm models and 1-D \mmm models (same as DPMM):
when $\omega = 1$, $z_i^1$ and $z_i^2$ will always be equal and
hence it simply boils down to a single DPMM; when $\omega = 0$, the equation above 
decomposes into two distributions similarly to Eq.~\eqref{eq:dp_CRP} for $z_i^1$ and $z_i^2$ respectively. 




We can use
 the algorithm of Gibbs sampling with auxiliary parameters~\cite{neal2000markov} to
sample $\boldsymbol z^1, \boldsymbol z^2, \Theta^1, \Theta^2$ from their posterior distributions: 
\begin{packed_enum}
\item For $i=1,\ldots,n$: sample $z_i^1$ and $z_i^2$ according to Eq.~\eqref{eq:dp2_CRP} multiplying
$f(x_i|\theta^1_{z_i^1}, \theta^1_{z_i^1})$;
\item For $l=1,2$: sample $\theta^l_k$ with the probability given by $G^l(\theta^l_k)
\prod_{i:z_i^l=k} f(x_i|\theta^1_{z_i^1},\theta^2_{z_i^2})$.
\end{packed_enum}
Here $f(\cdot|\theta^1,\theta^2)$ is the density function of distribution $F(\theta^1,\theta^2)$.


Implementations of the second step differ depending on which $F$ and $G^{1:2}$ are used. For example,
finding conjugate priors for exponential family distributions is easy for \mmm, but it is not so
when $G^{1:2}$ are both Dirichlet distributions because they are no longer the conjugate priors for
a multinomial distribution $F$.  In such cases, other methods based on sampling such as
Metropolis-Hastings \cite{neal2000markov} or Gibbs sampling~\cite{griffiths2004finding,HDP} could be
used
depending on the distribution. 
In this paper, we will present a concrete example with $F$ as a normal distribution and $G$ is its conjugate prior in Section~\ref{sec:mog_exp}. 

\vspace*{\subsectionReduceTop}
\subsection{Finite $M^3$ Models for Topic Modeling}
\vspace*{\subsectionReduceBot}
Latent Dirichlet allocation (LDA) 
employs a hierarchical finite mixture model to
describe the generative process of a document: first, a topic proportion $\bpi$ over $K$ topics is drawn from a
symmetric Dirichlet distribution with prior $\alpha$; then a topic $z_i$ is chosen
for each word
$x_i$, and $x_i$ is drawn from $\theta_{z_i}$, a multinomial distribution over the whole vocabulary
of size $V$. 
Thus, LDA allows  words from the same document share similar topic
distributions while documents share finite topics. 
However in many real-world datasets, words could be generated  from 
several different \emph{types} of topics.
In this section, we model each type of topic as a dimension in our $M^3$ model.



We define our 2-D finite \mmm model as follows (shown in Fig.~\ref{fig:models}d): we assume two independent topic spaces and a word is drawn from a
topic synthesized by two topics---one from each topic space---with the probability of
\begin{equation}
p(x|\theta^1_{z^1},\theta^2_{z^2}) = 
\frac{1+\omega}{2} \theta^1_{z^1,x} +\frac{1-\omega}{2} \theta^2_{z^2,x}
\end{equation}
where $\omega \in [0,1]$ tunes the weight of the two topic models in forming a new topic. It serves similar
purpose as the $\omega$ in Eq.~\eqref{eq:dp2_CRP}: the 2-D finite \mmm model degenerates to the classic LDA
when $\omega=1$.

Same as other topic models, the goal of applying finite \mmm model to corpora is density estimation, i.e.,
to maximize the likelihood of the test documents. After integrating out $\bpi^{1:2}$ and $\boldsymbol
z^{1:2}$, we obtain the likelihood of a document $\mathbf w=(x_1,\ldots,x_N)$ conditioned on the
model as,  
\begin{eqnarray}
p(\mathbf w | \alpha^{1:2}, \Theta^{1:2}) &\propto&
    \int\int \left(\prod_{i=1}^{K^1} (\pi^1_i)^{\alpha^1-1}\right)
            \left(\prod_{i=1}^{K^2} (\pi^2_i)^{\alpha^2-1}\right)\times \cr
    && \prod_{i=1}^N \sum_{z^1}^{K^1} \sum_{z^2}^{K^2} 
\pi^1_{z^1}\pi^2_{z^2} \left(\frac{1+\omega}{2} \theta^1_{z^1,x_i} + \frac{1-\omega}{2} \theta^2_{z^2,x_i}\right)
             d\boldsymbol\pi^1 d\boldsymbol\pi^2 .
\end{eqnarray}

This distribution is intractable to compute in general. We therefore approximate it using a
variational inference similar to the one used in LDA~\cite{LDA}.  

\noindent\textbf{Variational Inference.}
Following the classic LDA method, we use the variational distribution, 
$q(\bpi^{1:2}, \mathbf z^{1:2} | \bgamma^{1:2}, \phi^{1:2}) = 
    q(\bpi^1|\bgamma^1) q(\bpi^2|\bgamma^2) \prod_{n=1}^N q(z^1_n|\phi^1_n)q(z^2_n|\phi^2_n)$,
as an approximation to the true posterior distribution $p(\bpi^{1:2}, \boldsymbol z^{1:2} |\mathbf
w,\alpha^{1:2},\Theta^{1:2})$. 
The difference between the two is quantified by the KL divergence:
\begin{eqnarray}
D(q||p) 
    &=& \log p(\mathbf w|\alpha^{1:2},\Theta^{1:2}) - \mathcal
    L(\bgamma^{1:2},\phi^{1:2};\alpha^{1:2},\Theta^{1:2}).
\end{eqnarray}
Since KL divergence is always non-negative, $\mathcal L$ above is the lower bound of 
$p(\mathbf w|\alpha^{1:2},\beta^{1:2})$. Therefore,  
our goal is to maximize $\mathcal L$ so that the likelihood
$p(\mathbf w|\alpha^{1:2},\beta^{1:2})$ can be large as well.    
During inference, the goal is to optimize $\mathcal L$ 
with respect to $\phi^{1:2}$ and $\bgamma^{1:2}$ for
each document. This is similar to LDA and thus we provide details only in the supplementary
material. During training, given $D$ documents, our goal is to find the model's parameters 
that maximize $\mathcal L$. We solve it by iteratively
inferring $(\phi_d^{1:2},\bgamma_d^{1:2})$ for each document $\mathbf w_d$ and estimating
$\alpha^{1:2}, \Theta^{1:2}$ and $\omega$ given the rest. The step of parameter estimation is more
challenging than the classic LDA due to the entanglement of the two topics and 
the sharing parameter $\omega$.

\noindent\textbf{Parameter Estimation.} 
When the variational distribution is fixed, the terms involving $\alpha^{1:2}$ in $\mathcal L$ are as follows. Here, $\Gamma(\cdot)$ is the Gamma function, and $\Psi(\cdot)$ is the digamma function.
\begin{eqnarray}
\mathcal L_{\alpha^{1:2}} =
    \sum_{t=1}^2 \left( \log \Gamma(K^t \alpha^t)-K^t
    \log\Gamma(\alpha^t)+(\alpha^t-1)\sum_{i=1}^{K^t} \Big(\Psi(\gamma_i^t)-\Psi\Big(\sum_{j=1}^{K^t}\gamma_j^t \Big) \Big) \right).
\end{eqnarray}
Since $\alpha^1$ and $\alpha^2$ are independent to
each other and to $\omega$ and $\Theta^{1:2}$ 
as well. 
We can update them separately, similar to  LDA. However, the terms involving $\Theta^{1:2}$ and
$\omega$ in $\mathcal L$ are,
\begin{eqnarray}
\mathcal L_{\Theta^{1:2},\omega} = \sum_{d=1}^M\sum_{n=1}^{N_d}\sum_{i=1}^{K^1}\sum_{j=1}^{K^2} \phi^1_{dni}\phi^2_{dnj}\log
    \left(\frac{1+\omega}{2} \theta^1_{i,\mathbf w_{dn}}+\frac{1-\omega}{2}\theta^2_{j,\mathbf
    w_{dn}}\right).
\end{eqnarray}
Any derivative of this would have terms containing $2/((1+\omega) \theta^1_{i,\mathbf
w_{dn}}+(1-\omega)\theta^2_{j,\mathbf w_{dn}})$ in the inner-most summation, making it hard to obtain closed-form
expressions.
We instead convert the problem into an unconstrained problem, by defining the new objective
function: 
\begin{equation}
{\rm minimize}_{\Theta^1,\Theta^2,\omega} \qquad
    - \mathcal L_{\Theta^{1:2},\omega}  + 
    \frac{1}{2}\sum_{i=1}^{K^1} \lambda_i \Big(\sum_{j=1}^V \theta^1_{ij} -1 \Big)^2
    +\frac{1}{2}\sum_{i=1}^{K^2} \eta_i \Big(\sum_{j=1}^V \theta^1_{ij} -2 \Big)^2,
    \label{eq:optbeta}
\end{equation}
where \{$\lambda_i,\eta_i$\} impose a positive penalty for violating the constraint. Each
\{$\lambda_i,\eta_i$\} is initialized with a small value and gradually increased when the corresponding
constraint is violated.
Given the weights, optimal $\Theta^{1:2}$ and $\omega$ are computed by the limited-memory BFGS
algorithm, a standard quasi-Newton method. In practice, the penalties are only updated a few
times before it converges.



\vspace*{\subsectionReduceTop}
\subsection{Hybrid \mmm Models}
\vspace*{\subsectionReduceBot}
There are many cases when one mixture model has a fixed number of components while the other one
does not. For example, in the task of scene understanding, we can model object type and its pose
separately. While objects can appear in countless poses, it is reasonable to assume a finite set of
object categories. Therefore, we form our hybrid \mmm model as a combination of DPMM
and a finite mixture model. We update the assignments $z^1_i$ and $z^2_i$ in turn. 
We use standard DP to sample $z^1_i$ given $z^2_i$,  
and use maximum likelihood estimation to update $z^2_i$ given a set of sampled $z^1_i$.



 \vspace*{\sectionReduceTop}
\section{Applications and Experimental Results}\label{sec:experiments}
\vspace*{\sectionReduceBot}

In this section, we first illustrate how our \mmm models behaves on the task of Gaussian mixture
modeling. Then we evaluate our finite \mmm model on the task of document modeling, 
on four different datasets and against three baselines. Finally, we apply our hybrid \mmm model 
to the task of estimating object arrangements in human environments.   

\begin{figure}[!t]
\begin{minipage}[!t]{.45\linewidth}
\centering
\includegraphics[height = .30\linewidth, width=.32\linewidth]{mog_data3.pdf}
\includegraphics[height = .30\linewidth, width=.32\linewidth]{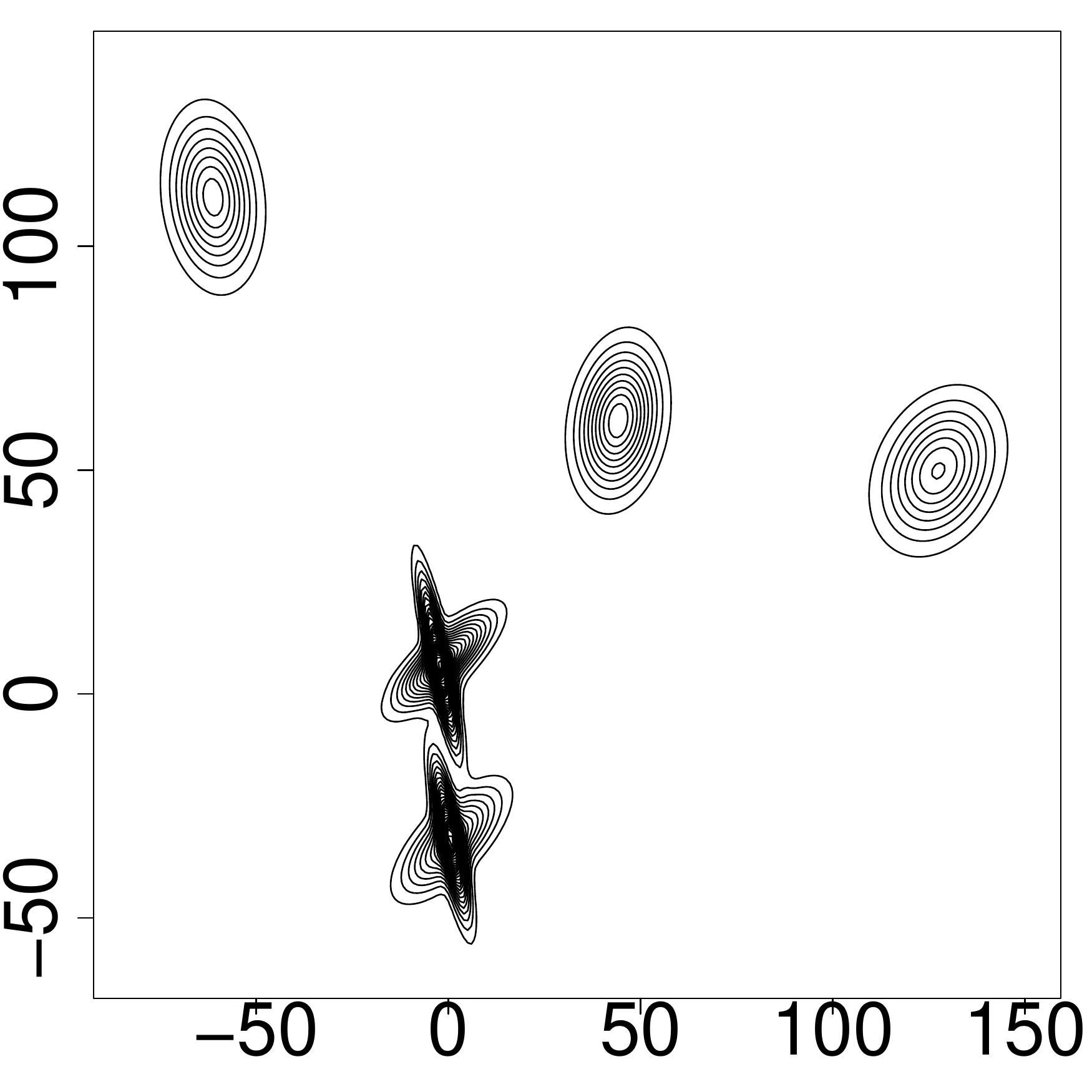}
\includegraphics[height = .30\linewidth, width=.32\linewidth]{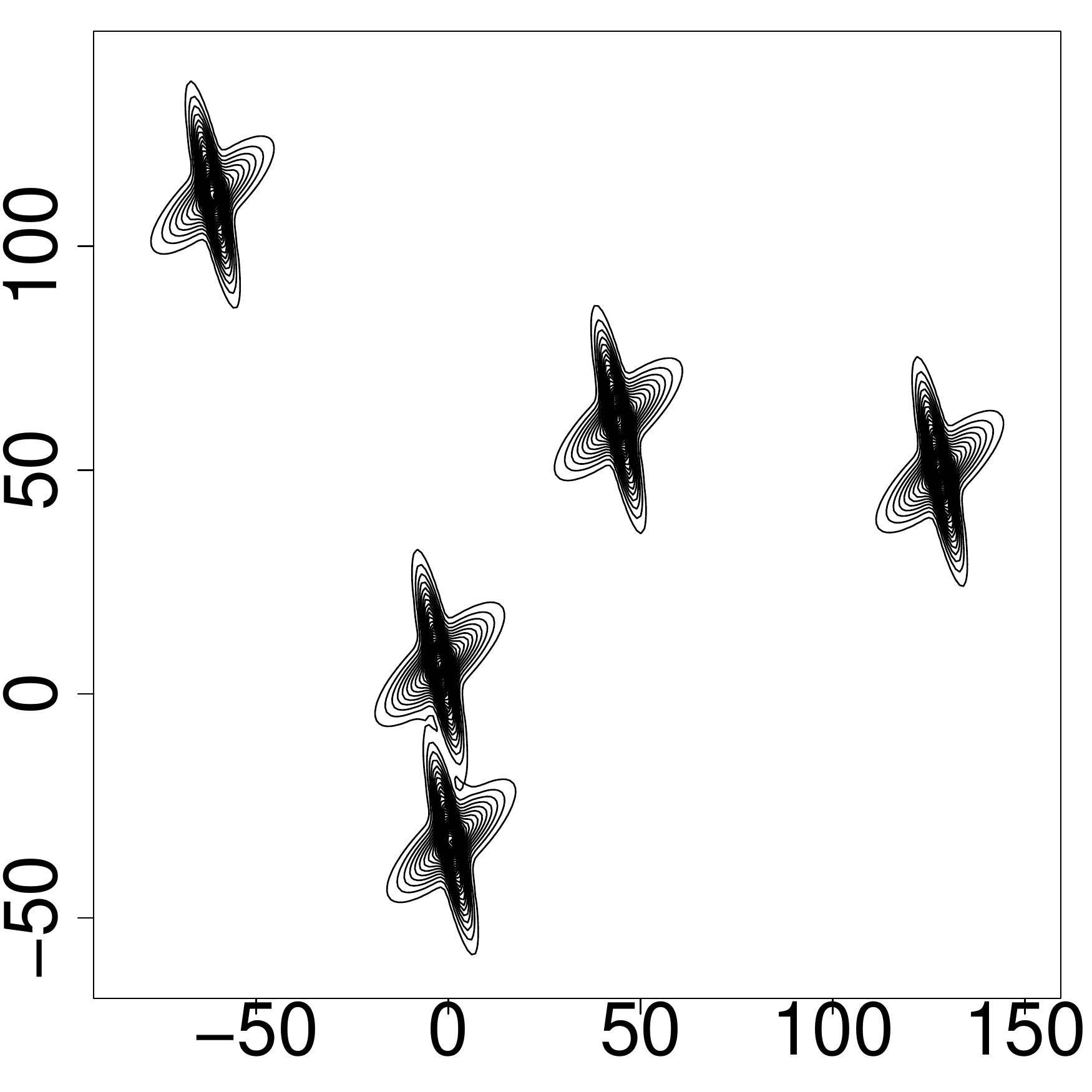}\\
\includegraphics[height = .30\linewidth, width=.32\linewidth]{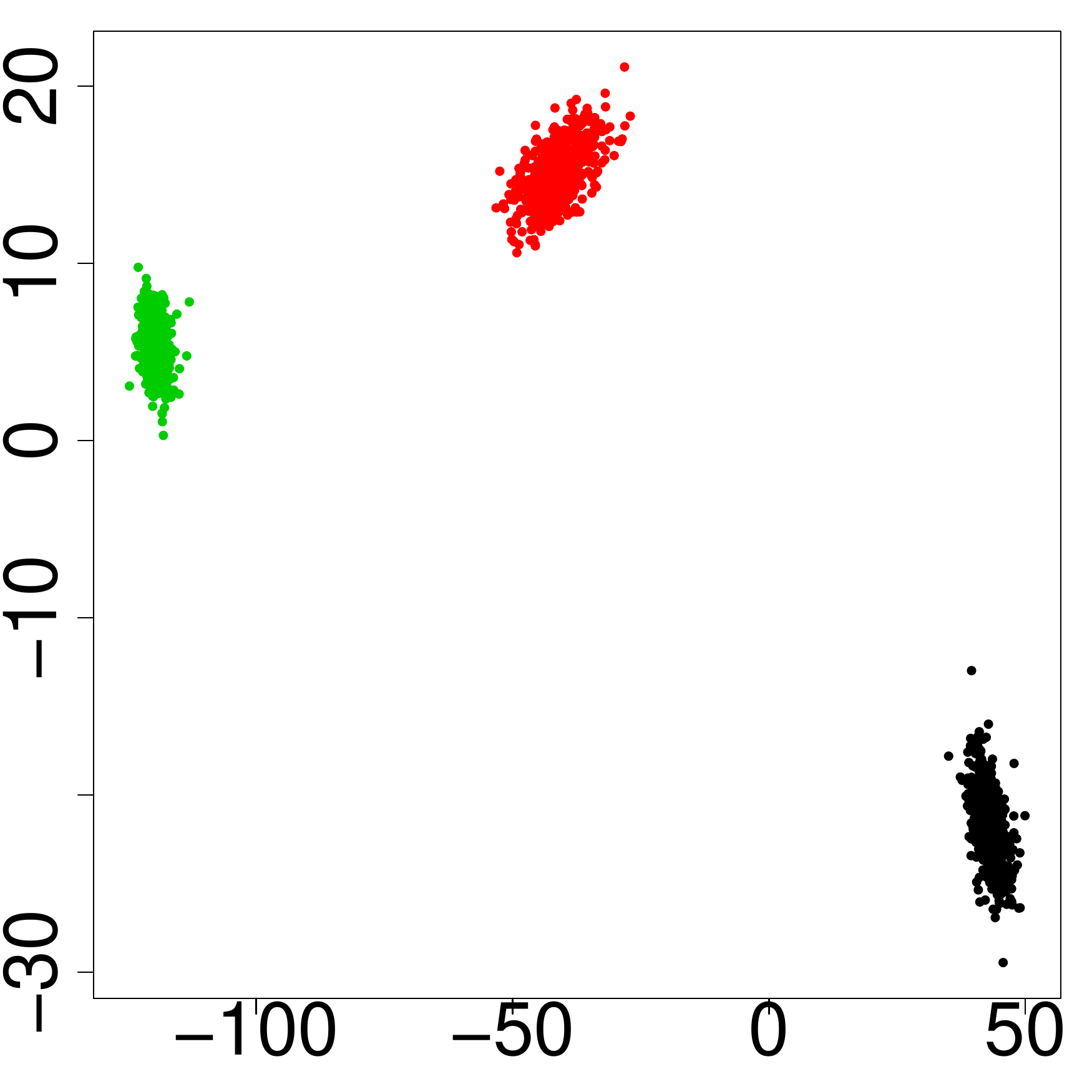}
\includegraphics[height = .30\linewidth, width=.32\linewidth]{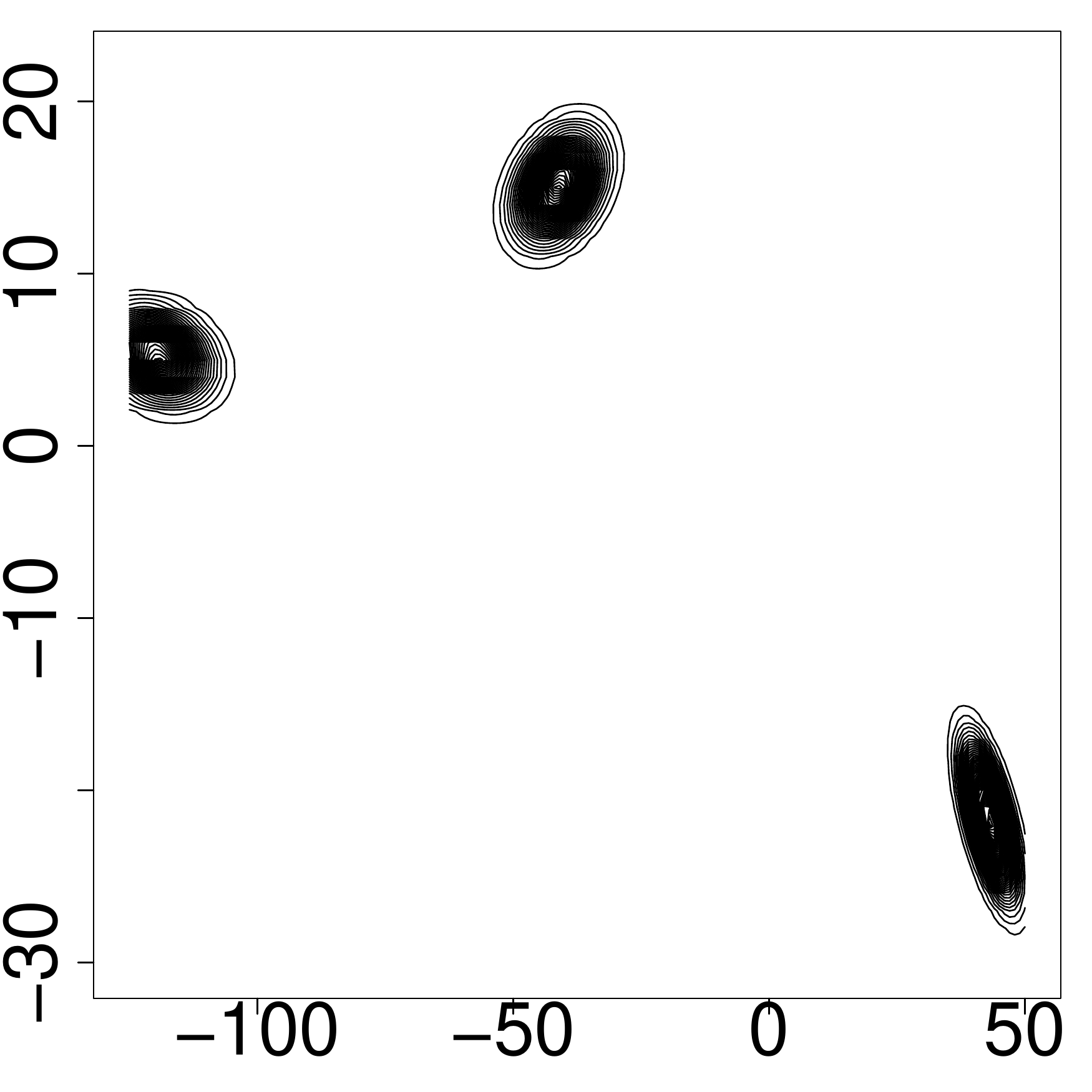}
\includegraphics[height = .30\linewidth, width=.32\linewidth]{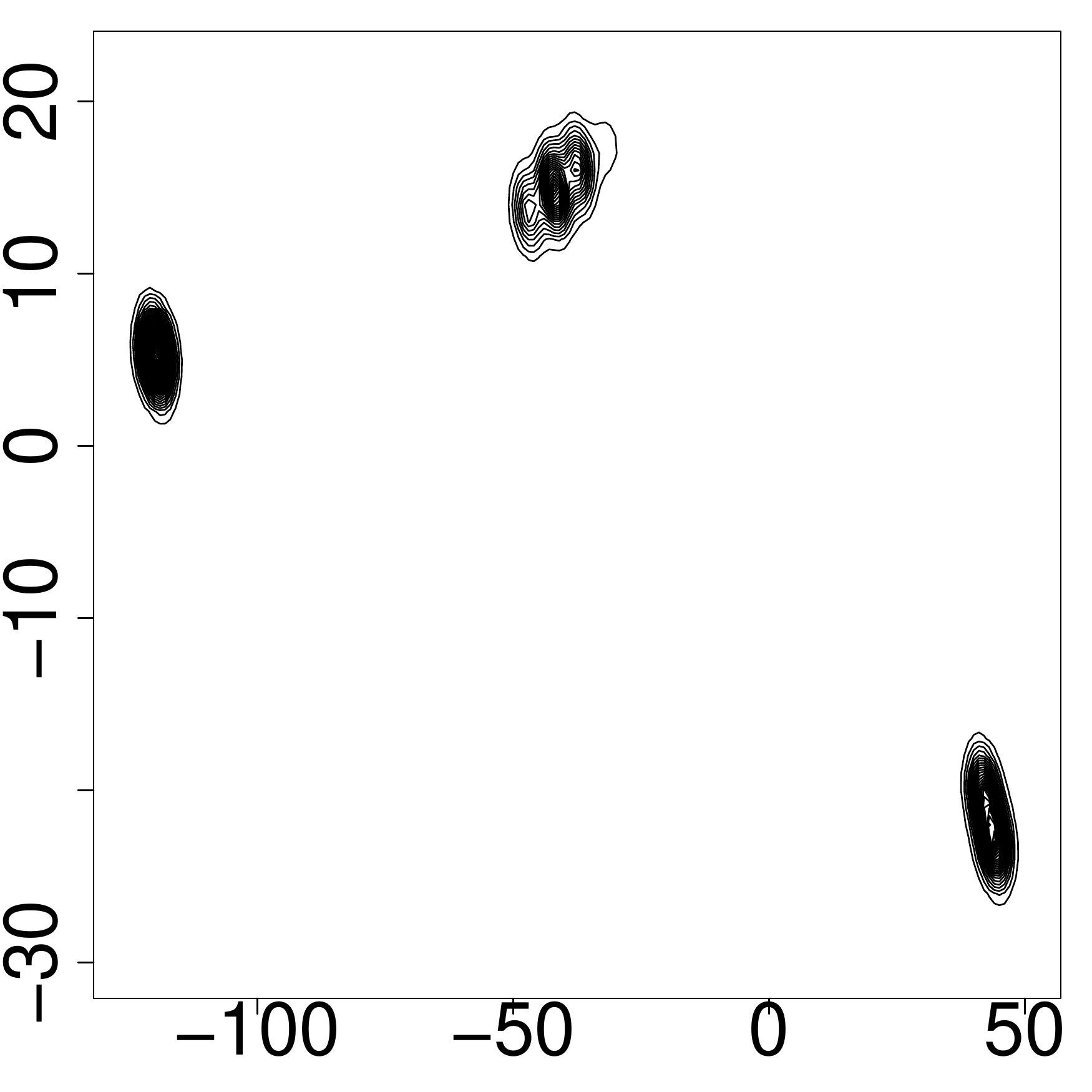}
\vskip -.1in
\caption{Given data generated from a mixture of Gaussian distributions (left), the  
density estimation obtained by the standard DPMM (middle) and our infinite \mmm model (right). 
Bottom row shows that even when there are no shared parameters, our model performs as well 
as the DPMM.}
\label{fig:gaussian}
\end{minipage}
\hskip .25in
\begin{minipage}[!t]{.53\linewidth}
\centering
\includegraphics[width=.49\linewidth]{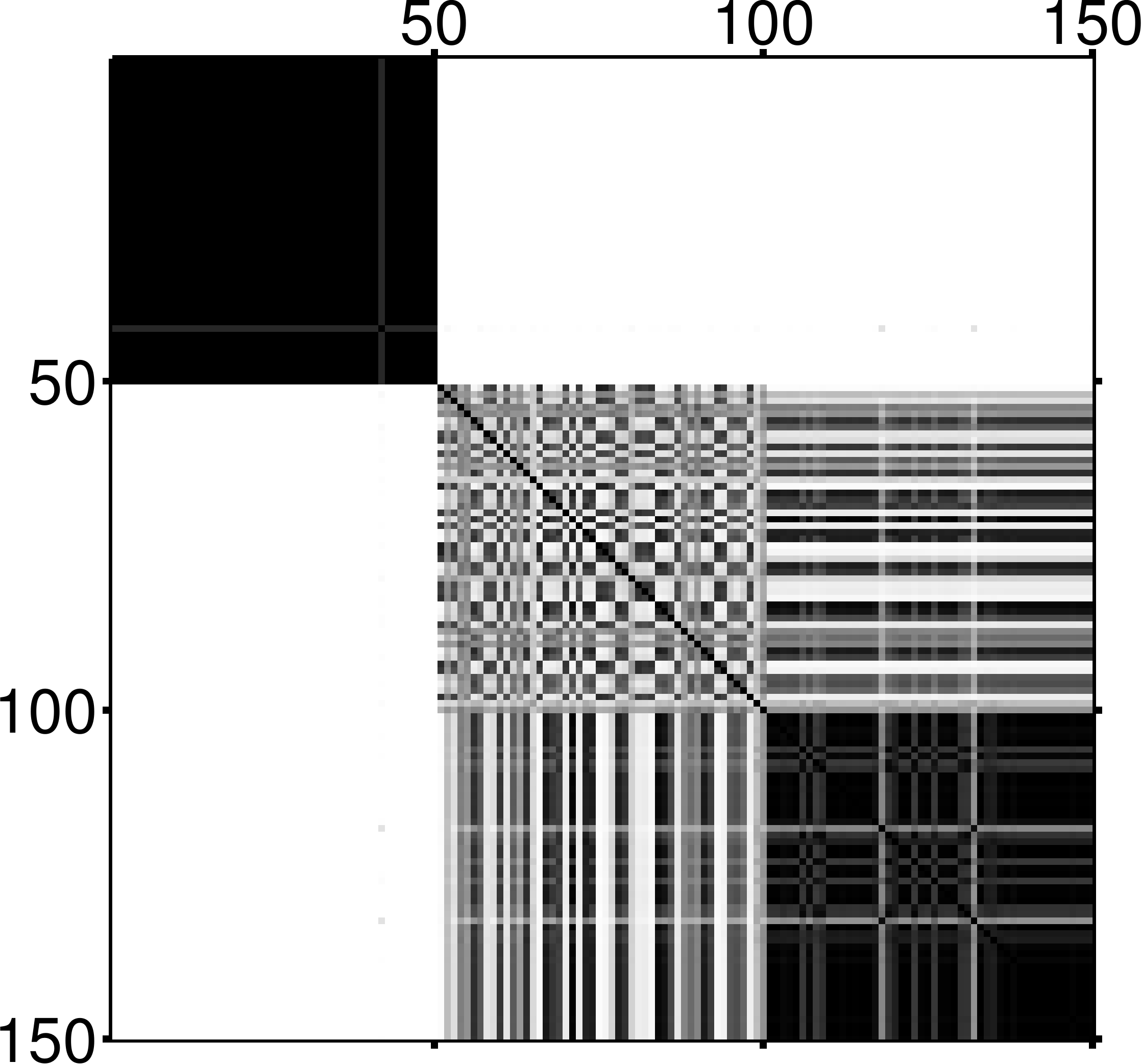}
\includegraphics[width=.49\linewidth]{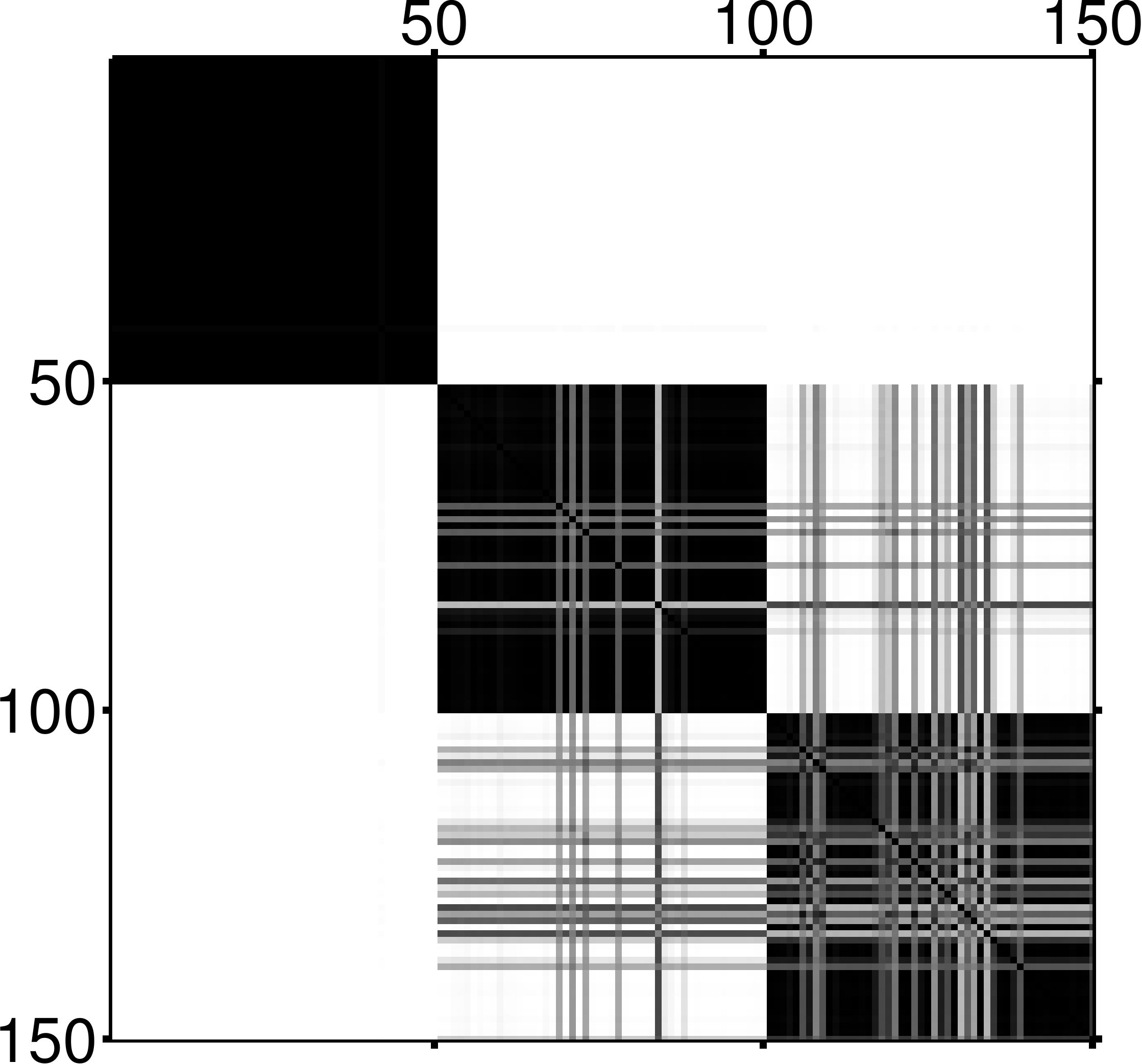}
\caption{Confusion matrices for \textbf{Dataset-0} obtained by the DPMM (left) and infinite \mmm
model (right). The intensity of each pixel represents the percentage of two flowers are grouped into
one cluster. The ground-truth is a $3\times 3$ block-diagonal matrix. 
}\label{fig:iris}
\end{minipage}
\end{figure}

\vspace*{\subsectionReduceTop}
\subsection{Gaussian Mixture Model}\label{sec:mog_exp}
\vspace*{\subsectionReduceBot}
In the classic Gaussian mixture model, data points are drawn from a set of different Gaussian distributions, i.e. 
$x_i|z_i, (\mu_1,\Sigma_1),\ldots, (\mu_K,\Sigma_K) \sim N(\mu_{z_i}, \Sigma_{z_i})$, whereas our
2D \mmm model uses two mixture models for means and covariances respectively, and draws data from 
$$x_i|z_i^1,z_i^2,\mu_1,\ldots,\mu_{K^1}, \Sigma_1,\ldots, \Sigma_{K^2} \sim N(\mu_{z_i^1},
\Sigma_{z_i^2}).$$
We use conjugate prior for $\mu$ and $\Sigma$ (Gaussian and inverse Wishart distribution respectively), with same hyperparameters in
both algorithms. 


We created a synthetic dataset for evaluating the results in terms of density estimation. 
For our model, it is $x \sim \sum_c\sum_d p(c,d|\boldsymbol z^1, \boldsymbol
z^2) \mathcal N(\mu_c, \Sigma_d)$ averaged over 1000 samples. From the contours shown in
Fig.~\ref{fig:gaussian}, we can see that our method successfully identifies correct clusters in both
sharing and non-sharing cases. The averaged normalized mutual information (NMI) for our model is
$0.75$ and $0.96$ compared to $0.66$ and $0.97$ of the DPMM. 

We also tested it on the Iris dataset (\textbf{Dateset-0}) containing 150 flowers from three species with four features
for each sample.\footnote{http://archive.ics.uci.edu/ml/datasets/Iris}
The confusion matrices in Fig.~\ref{fig:iris} show that
while both methods can correctly find the first species, DPMM is  confused about the last two
species. NMI of \mmm model is 0.72 versus 0.67 for the DPMM. 



\begin{figure}
\vskip -.1in
\centering
\includegraphics[width=.6\linewidth]{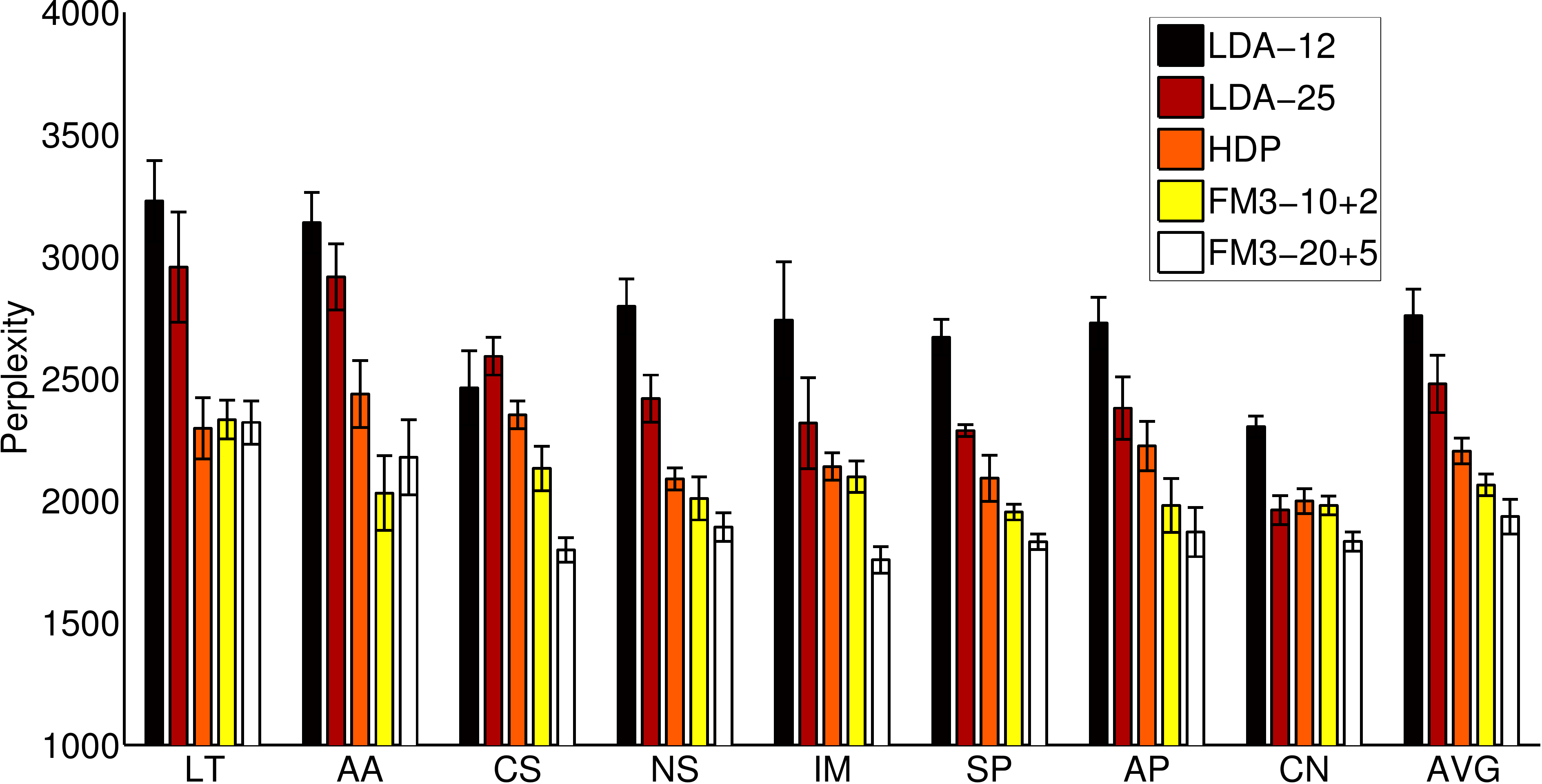}
\includegraphics[width=.39\linewidth]{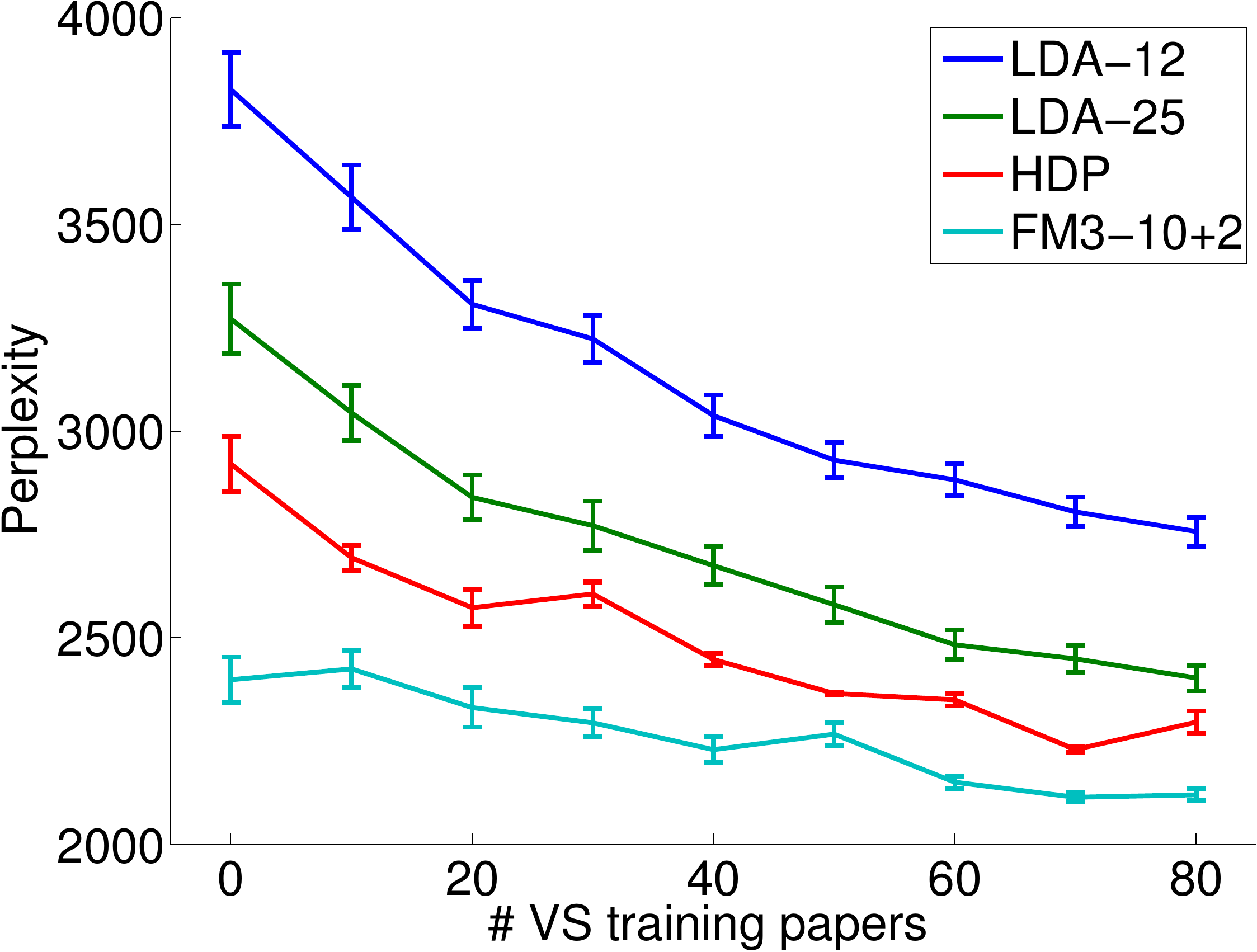}
\vspace*{\captionReduceTop}
\caption{\textbf{Results on Dataset-1.} Perplexity of mixing VS and other 8 sections (left) and the average perplexity when
changing the number of training documents from VS (right). The error bars are one standard error.}
\vspace*{\captionReduceBot}
\label{fig:nips}
\end{figure}

\vspace*{\subsectionReduceTop}
\subsection{Document Topic Modeling}
\vspace*{\subsectionReduceBot}
We test our finite \mmm model (tagged `FM3' in the figures) on four document corpora: 
\textbf{Dataset-1} contains processed NIPS 1-12 proceedings with 1447 papers organized into 9
sections and 5270 words after removing words appeared more than 4000 times
or fewer than 50 times;\footnote{\label{nips-corpus}http://www.gatsby.ucl.ac.uk/$\sim$ywteh/research/data.html}  
\textbf{Dataset-2} includes randomly selected 1000 documents from the 20 newsgroups with a total of
1498 words after removing stop-words and words in fewer than 5 documents;\footnote{http://people.csail.mit.edu/jrennie/20Newsgroups/}
\textbf{Dataset-3} selects 1000 encyclopedia articles with 1200 words;\footnote{http://www.cs.nyu.edu/$\sim$roweis/data.html}
\textbf{Dataset-4} takes 500 articles from Psychological Review with 1244 words.\footnote{http://psiexp.ss.uci.edu/research/programs\_data/toolbox.htm}
All the following results are based on 5 runs or 5-fold cross validation. 
Experiments on the first two datasets are in the same setup as in~\cite{HDP} and
\cite{focussedTopicModelsNIPSws} respectively. 

To investigate how well our model can learn general topics and section-specific topics, we train on
80 articles from the VS (vision science) section and 80 articles from one of the other 8 sections. We
test on the rest 47 VS papers. We use the perplexity~\cite{LDA} of the on-hold documents to evaluate
the learned topic model:
$\textstyle
\text{perplexity}(\textbf w_1, \ldots, \textbf w_D) = 
\exp (- (\sum_{d=1}^D \log p(\textbf w_d))/\sum_{d=1}^D N^d)
$.
A lower perplexity indicates higher likelihood of the test data and thus better performance. 

Fig.~\ref{fig:nips}-left shows the perplexity obtained by LDA, HDP and our method. 
In the comparison with LDA, we set the LDA's topic number $K$ equal to the total sum of topic
numbers of \mmm model $K^1+K^2$, so that the two models have the same number of parameters.  
We see that our method performs significantly better than LDA across all eight sections for both 12
and 25 topics. This is due to that \mmm model has effectively have more topics than LDA. 
Such trends hold for different values of $K$, $K^1$ and $K^2$.

In presented results, we had set the second dimension of our \mmm model to have 
only a few topics ($K^2 = 2$ or $5$). This enforces all documents from different sections 
have to `share' them.  
Our method and HDP both outperform than LDA, showing that the ability of having shared topics is
helpful. However, HDP does so in a hierarchy so that a sub-tree share similar topic proportions. It
however does not reduce the number of topics needed to model. 
In fact, the number of topics used in HDP is around 55, far more than our 12 topics.  

In another experiment on Dataset-1, we change the number of training documents from 
VS from 0 to 80, but
always test on the rest 47 VS documents. When the number is small, the domain of 
the training and
test dataset would be different and thus can be used to test the transfer of topic learning. 
Fig.~\ref{fig:nips}-right shows the perplexity, averaged
over all sections, with respect to different training documents. We can see that the performance of
LDA largely depends on the number of VS papers, while the change in the perplexity of HDP and our
model is less significant. Our finite \mmm method not only beats all the baselines but also gives
the most consistent results in all cases. This demonstrates that 1) our model can learn the common topics
of two different sections, and 2) it is less sensitive to having a small training set since the
multi-dimensional membership effectively allows more documents to be used for estimating the parameters. 

\begin{figure}[t!]
\vskip -.15in
\centering
\begin{minipage}{.5\linewidth}
\includegraphics[width=.97\linewidth]{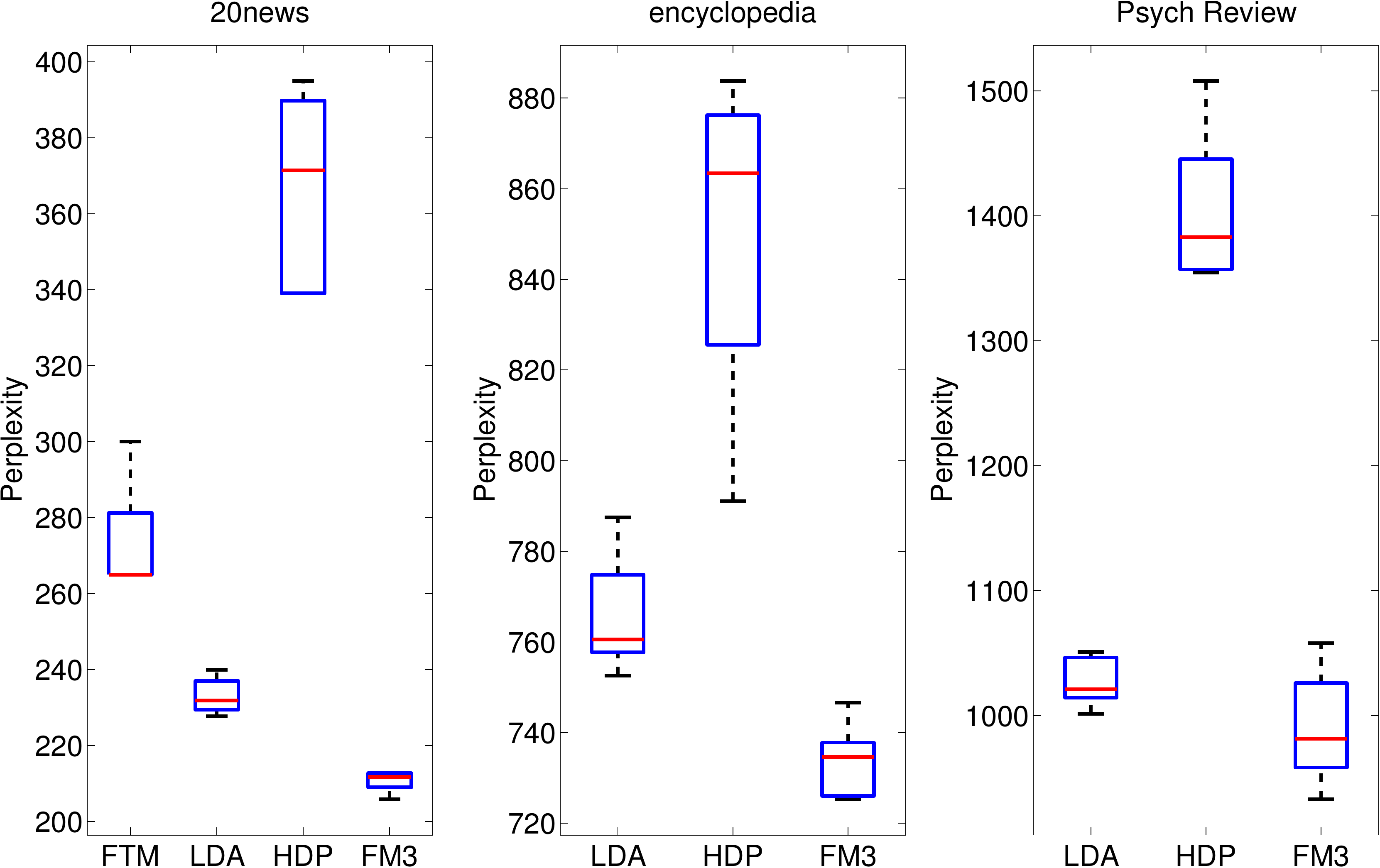}
\caption{Results on \textbf{Dataset2-4}
, performed by FTM (reported in~\cite{focussedTopicModelsNIPSws}), LDA, HDP and our finite
\mmm model. For LDA and FM3, we only report the best performance among different number of topics.}
\label{fig:20news}
\end{minipage}
\quad
\begin{minipage}{.45\linewidth}
\centering
{\tiny
\begin{tabular}{cccc}
\hline
 VS+AP & VS+CN & VS+SP\\
\hline
   parameter	&    response	& force\\
   measure	&    murray	& digits\\
   subthreshold	&    receptive	&    bifurcation\\
   versions	&    lower	&    brain\\
   tuned	&    type	&    dimension\\
   bound	&    statistical	&    realized\\
   obtain	&    bifurcation	&    electrodes\\
\hline
   work	&    work	&    response\\
   algorithms	&    section	&    cells\\
   analysis	&    features	&    algorithms\\
   images	&    neuron	&    form\\
   achieved	&    problems	&    rate\\
   form	&    obtained	&    low\\
   local	&    images	&    local\\
\hline
\end{tabular}
}
\vspace*{\captionReduceTop}
\caption{Topics of VS combined with three other sections found by our finite \mmm model:
top seven words (ranked by weight) of one topic from each dimension ($K^1=10, K^2=2$) is listed. 
Topics from first dimension are section-specific and different from each other while the second
dimension contains popular terms in NIPS and the topics in it do not change much.
}
\label{fig:topics}
\end{minipage}
\vskip -.25in
\end{figure}

We also test on the other three datasets and the results are shown in Fig.~\ref{fig:20news}. 
Compared to the baselines, our \mmm model obtains the lowest perplexity and
demonstrates its robustness in different scenarios.   

In order to explore what orthogonal topics our \mmm model  discovered, we list one topic from each
dimension in Fig.~\ref{fig:topics}. Topics from the first row are quite different from each other
containing some keywords for specific sections, such as `digits' for the SP (speech and signal
processing) while the topics in the bottom row are mostly from popular words in NIPS such as `work'
and `algorithms'.  This indeed reflects that \mmm represents topics parsimoniously.

In \mmm models, setting $K^1$ and $K^2$ would affect the performance. Similar to LDA, the optimal
value for the number of topics ($K^1+K^2$) varies with the size and heterogeneity of the corpus,
and may to try different values. The ratio $K^1/K^2$ is  interesting---in most datasets
we found that an asymmetric value performs better, e.g., the result of setting $K^1=K^2=10$
is worse than $K^1=20,K^2=5$.

\vspace*{\subsectionReduceTop}
\subsection{Object Arrangement}
\vspace*{\subsectionReduceBot}

\begin{wraptable}[7]{r}{.37\linewidth}
\vskip -.3in
\centering
\caption{Results of learning object arrangements evaluated by the difference in location and height
(in meter).}\label{tbl:arrange}
\vskip -.1in
{\scriptsize
\begin{tabular}{@{}l|cc|cc@{}}
\hline
& \multicolumn{2}{c|}{new object} & \multicolumn{2}{c}{empty room} \\
\cline{2-5}
& location & height & location & height\\
\hline
FMM & 1.59 & 0.16 & 1.74 & 0.20 \\ 
DP & 1.65 & 0.11 & 2.01 & 0.28 \\ 
\hline
\mmm & \textbf{1.44} & \textbf{0.09} & \textbf{1.63} & \textbf{0.11}\\
\hline
\end{tabular}
}
\end{wraptable}

We finally considered the task of learning object arrangement in human scenes, using the dataset
in~\cite{JiangICML}. It contains 3D models for 20 scenes such as living rooms, kitchens and offices,
and 19 different object categories.
Each room was manually labeled with arrangements of multiple objects.  
We considered two learning scenarios: placing new objects that are not in the test room and placing in an empty room. 
We performed 5-fold cross validation on 20 rooms and evaluated the
predicted arrangements based on the location difference and height difference from the labels.
We compared our hybrid \mmm model (with one DPMM for human poses and one finite mixture model for
object affordances) against both a finite mixture model and a single DPM. Results are shown in
Table~\ref{tbl:arrange}. Our method not only predicts arrangements closer to the ground truth but
also places relevant but different type of objects together due to it allows objects with different
affordances share the same human pose.

\vspace*{\sectionReduceTop}
\section{Conclusion}
\vspace*{\sectionReduceBot}
In this paper, we presented the multidimensional membership mixture (\mmm) models which consists of
multiple independent mixture models. Each data point is generated from a set of mixture components
jointly, designated by its multidimensional membership.
We derived three instantiations of \mmm models---infinite, finite and hybrid \mmm. The infinite \mmm
model uses multiple Dirichlet processes as the prior of memberships while the finite \mmm is built
upon two LDAs. In both models, we introduced a tunable sharing parameter to increase its robustness
in both sharing and no-sharing situations. The challenge in inference is addressed by Gibbs sampling
and variational inference.
We applied \mmm models on topic modeling. Compared to the baselines, our model demonstrated its ability in achieving better performance
with fewer topics and in learning orthogonal topics. We also verified our model in the 
application of learning object arrangements. 

\setlength{\bibsep}{0pt}
{\footnotesize
 \bibliographystyle{abbrv}
\bibliography{nips12ref}
}

\end{document}